\documentclass[11pt,a4paper]{article}
\usepackage[hyperref]{acl2020}
\usepackage{times}
\usepackage{latexsym}

\usepackage{microtype}

\usepackage{url}
\usepackage{multirow,multicol}
\usepackage{amsmath}
\usepackage{tabularx}
\usepackage{makecell}
\usepackage{subfig}

\usepackage{helvet}
\usepackage{courier}
\usepackage{bm}
\usepackage{amsmath}
\usepackage{mdwlist}
\usepackage[boxed]{algorithm2e}
\usepackage[noend]{algpseudocode}
\usepackage{relsize}
\usepackage{amsfonts}
\usepackage{graphicx}
\usepackage{caption}
\usepackage{hhline}
\usepackage{color}
\usepackage{colortbl}
\usepackage{booktabs}
\usepackage{CJKutf8}
\usepackage{multirow}
\usepackage{arydshln}

\aclfinalcopy % Uncomment this line for the final submission
\def\aclpaperid{828} %  Enter the acl Paper ID here

\title{Simplify the Usage of Lexicon in Chinese NER}

\author{Ruotian Ma$^1$\thanks{{ }{ }Equal contribution.}{ }{ }, Minlong Peng$^{1*}$, Qi Zhang$^{1,3}$, Zhongyu Wei$^{2,3}$, Xuanjing Huang${^1}$\\
$^1$Shanghai Key Laboratory of Intelligent Information Processing, \\
School of Computer Science, Fudan University\\
$^2$School of Data Science, Fudan University\\
$^3$Research Institute of Intelligent and Complex Systems,
Fudan University\\
\{rtma19,mlpeng16,qz,zywei,xjhuang\}@fudan.edu.cn}

\date{}

\begin{document}
\begin{CJK}{UTF8}{gbsn}
\maketitle
\begin{abstract}
Recently, many works have tried to augment the performance of Chinese named entity recognition (NER) using word lexicons. As a representative, Lattice-LSTM \cite{zhang2018chinese} has achieved new benchmark results on several public Chinese NER datasets. However, Lattice-LSTM has a complex model architecture. This limits its application in many industrial areas where real-time NER responses are needed. 

In this work, we propose a simple but effective method for incorporating the word lexicon into the character representations. This method avoids designing a complicated sequence modeling architecture, and for any neural NER model, it requires only subtle adjustment of the character representation layer to introduce the lexicon information. Experimental studies on four benchmark Chinese NER datasets show that our method achieves an inference speed up to 6.15 times faster than those of state-of-the-art methods, along with a better performance. The experimental results also show that the proposed method can be easily incorporated with pre-trained models like BERT. \footnote{The source code of this paper is publicly available at \url{https://github.com/v-mipeng/LexiconAugmentedNER}.}%It also shows that our method can apply to different neural architectures. 
\end{abstract}

\section{Introduction}

Named Entity Recognition (NER) is concerned with the identification of named entities, such as persons, locations, and organizations, in unstructured text. 
NER plays an important role in many downstream tasks, including knowledge base construction \cite{riedel2013relation}, information retrieval~\cite{chen2015event}, and question answering~\cite{diefenbach2018core}.
In languages where words are naturally separated (e.g., English), NER has been conventionally formulated as a sequence labeling problem, and the state-of-the-art results have been achieved using neural-network-based models \cite{huang2015bidirectional,chiu2016named,Liu18ner}.

Compared with NER in English, Chinese NER is more difficult since sentences in Chinese are not naturally segmented. Thus, a common practice for Chinese NER is to first perform word segmentation using an existing CWS system and then apply a word-level sequence labeling model to the segmented sentence ~\cite{yang2016combining,he2017unified}. However, it is inevitable that the CWS system will incorrectly segment query sentences. This will result in errors in the detection of entity boundary and the prediction of entity category in NER.
% Take the character sequence ``南京市 (Nanjing) / 长江大桥 (Yangtze River Bridge)" as an example, where ``/" indicates the gold segmentation result. If the sequence is segmented into ``南京 (Nanjing) / 市长 (mayor) / 江大桥 (Daqiao Jiang)", the word-based NER system is not able to correctly recognize ``南京市 (Nanjing)" and ``长江大桥 (Yangtze River Bridge)" as two location names. Instead, it is very likely to predict ``南京 (Nanjing)" as a location name and predict ``江大桥 (Daqiao Jiang)" as a person's name. 
Therefore, some approaches resort to performing Chinese NER directly at the character level, which has been empirically proven to be effective \cite{he2008chinese,liu2010chinese,li2014comparison, liu2019encoding, sui2019leverage, gui2019lexicon, ding2019neural}.

A drawback of the purely character-based NER method is that the word information is not fully exploited. With this consideration, \citeauthor{zhang2018chinese}, \shortcite{zhang2018chinese} proposed Lattice-LSTM for incorporating word lexicons into the character-based NER model. Moreover, rather than heuristically choosing a word for the character when it matches multiple words in the lexicon, the authors proposed to preserve all words that match the character, leaving the subsequent NER model to determine which word to apply. To realize this idea, they introduced an elaborate modification to the sequence modeling layer of the LSTM-CRF model \cite{huang2015bidirectional}. Experimental studies on four Chinese NER datasets have verified the effectiveness of Lattice-LSTM.

However, the model architecture of Lattice-LSTM is quite complicated. In order to introduce lexicon information, Lattice-LSTM adds several additional edges between nonadjacent characters in the input sequence, which significantly slows its training and inference speeds.
% and the implementation of Lattice-LSTM is too complicated to perform batch training and decoding. 
In addition, it is difficult to transfer the structure of Lattice-LSTM to other neural-network architectures (e.g., convolutional neural networks and transformers) that may be more suitable for some specific tasks.

% In order to introduce the lexicon information, Lattice-LSTM adds additional edges between nonadjacent characters of the input sequence. This seriously slows down its training and inference speed. In addition, the implementation of Lattice-LSTM is too complicated to perform batch training and inference, and to transfer the structure to other neural-network architectures (e.g., convolutional neural networks and transformers) that may be more suitable for some specific tasks.

In this work, we propose a simpler method to realize the idea of Lattice-LSTM, i.e., incorporating all the matched words for each character to a character-based NER model. The first principle of our model design is to achieve a fast inference speed. To this end, we propose to encode lexicon information in the character representations, and we design the encoding scheme to preserve as much of the lexicon matching results as possible. Compared with Lattice-LSTM, our method avoids the need for a complicated model architecture, is easier to implement, and can be quickly adapted to any appropriate neural NER model by adjusting the character representation layer. 
In addition, ablation studies show the superiority of our method in incorporating more complete and distinct lexicon information, as well as introducing a more effective word-weighting strategy. The contributions of this work can be summarized as follows:
\begin{itemize}
    \item We propose a simple but effective method for incorporating word lexicons into the character representations for Chinese NER.%, avoiding designing complicated model architecture.
    \item The proposed method is transferable to different sequence-labeling architectures and can be easily incorporated with pre-trained models like BERT \cite{devlin2018bert}.
    % \item Experiments on four public Chinese NER datasets show that, when implementing with a single-layer Bi-LSTM, the proposed method can achieve considerable improvement over Lattice-LSTM in both inference speed and sequence labeling performance.
\end{itemize}
We performed experiments on four public Chinese NER datasets.
The experimental results show that when implementing the sequence modeling layer with a single-layer Bi-LSTM, our
method achieves considerable improvements over the state-of-the-art methods in both inference speed and sequence labeling performance. 
% In addition, the experiments show that the proposed method can be easily incorporated with pre-trained models like BERT. 

\section{Background}
In this section, we introduce several previous works that influenced our work, including the Softword technique and Lattice-LSTM.
% A problem with the above character-based NER method is that it fails to exploit the word information. Therefore, several approaches have been proposed to incorporate word information for the character-based NER model.

\subsection{Softword Feature} 
The Softword technique was originally used for incorporating word segmentation information into downstream tasks \cite{zhao2008unsupervised,peng2016improving}. It augments the character representation with the embedding of its corresponding segmentation label: 
\begin{equation}
\bm{x}^c_j \leftarrow [\bm{x}^c_j;\bm{e}^{seg}(seg(c_j))].
\end{equation}
Here, $seg(c_j) \in \mathcal{Y}_{seg}$ denotes the segmentation label of the character $c_j$ predicted by the word segmentor, $\bm{e}^{seg}$ denotes the segmentation label embedding lookup table, and typically $\mathcal{Y}_{seg}=\{\text{B}, \text{M}, \text{E}, \text{S}\}$.
% where B, M, E indicates that the character lies in the beginning, middle, and end of a word, respectively, and S indicates that the character itself forms a single-character word. 

However, gold segmentation is not provided in most datasets, and segmentation results obtained by a segmenter can be incorrect. Therefore, segmentation errors will inevitably be introduced through this approach.

\subsection{Lattice-LSTM}
Lattice-LSTM designs to incorporate lexicon information into the character-based neural NER model. To achieve this purpose, lexicon matching is first performed on the input sentence. If the sub-sequence $\{c_i, \cdots, c_j\}$ of the sentence matches a word in the lexicon for $i < j$, a directed edge is added from $c_i$ to $c_j$. All lexicon matching results related to a character are preserved by allowing the character to be connected with multiple other characters. 
% Concretely, for a sentence $\{c_1, c_2, c_3, c_4, c_5\}$, if both of its sub-sequences $\{c_1, c_2, c_3, c_4\}$ and $\{c_2, c_3, c_4\}$ match a word of the lexicon, a directed edge from $c_1$ to $c_4$ and a directed edge from $c_2$ to $c_4$ will be added to the sentence.
\textit{Intrinsically, this practice converts the input form of a sentence from a chain into a graph.}

In a normal LSTM layer, the hidden state $\bm{h}_i$ and the memory cell $\bm{c}_i$ of each time step is updated by:
\begin{equation}
\bm{h}_i, \bm{c}_i = f(\bm{h}_{j-1}, \bm{c}_{j-1}, \bm{x}^c_j),
\end{equation}
However, in order to model the graph-based input, Lattice-LSTM introduces an elaborate modification to the normal LSTM. Specifically, let $s_{< *, j>}$ denote the list of sub-sequences of sentence $s$ that match the lexicon and end with $c_j$, $\bm{h}_{< *, j>}$ denote the corresponding hidden state list $\{\bm{h}_i, \forall s_{<i, j>} \in s_{< *, j>}\}$, and $\bm{c}_{< *, j>}$ denote the corresponding memory cell list $\{\bm{c}_i, \forall s_{<i, j>} \in s_{< *, j>}\}$. In Lattice-LSTM, the hidden state $\bm{h}_j$ and memory cell $\bm{c}_j$ of $c_j$ are now updated as follows:
\begin{equation}
\resizebox{0.9\columnwidth}{!}{$
\bm{h}_j, \bm{c}_j = f(\bm{h}_{j-1}, \bm{c}_{j-1}, \bm{x}^c_j, s_{<*, j>}, \bm{h}_{<*, j>}, \bm{c}_{<*, j>}),$}
\end{equation}
where $f$ is a simplified representation of the function used by Lattice-LSTM to perform memory update. 
% Note that, in the updating process, the inputs now contains current step character representation $\bm{x}_j^c$, last step hidden state $\bm{h}_{j-1}$ and memory cell $\bm{c}_{j-1}$, and lexicon matched sub-sequences $s_{<*, j>}$ and their corresponding hidden state and memory cell lists, $\bm{h}_{<*, j>}$ and $\bm{c}_{<*, j>}$. We refer you to the paper of Lattice-LSTM \cite{zhang2018chinese} for more detail of the implementation of $f$.

From our perspective, there are two main advantages to Lattice-LSTM. First, it preserves all the possible lexicon matching results that are related to a character, which helps avoid the error propagation problem introduced by heuristically choosing a single matching result for each character. Second, it introduces pre-trained word embeddings to the system, which greatly enhances its performance. 
% The disadvantage of Lattice-LSTM is that it converts the sentence input from a chain into a graph. This greatly increases its computational cost for sentence modeling. Therefore, our method aims to maintain the chained input form of the sentence while achieving the above two advantages of Lattice-LSTM.

However, efficiency problems exist in Lattice-LSTM. Compared with normal LSTM, Lattice-LSTM needs to additionally model $s_{<*, j>}$, $\bm{h}_{<*, j>}$, and $\bm{c}_{<*, j>}$ for memory update, which slows the training and inference speeds. Additionally, due to the complicated implementation of $f$, it is difficult for Lattice-LSTM to process multiple sentences in parallel (in the published implementation of Lattice-LSTM, the batch size was set to 1). These problems limit its application in some industrial areas where real-time NER responses are needed.

% \begin{figure}
% \centering
% \includegraphics[width=1\columnwidth]{model2.png}
% \caption{The overall architecture of the proposed method. }
% \label{fig:model2}
% \end{figure}

\section{Approach}
\begin{figure}
\centering
\includegraphics[width=1\columnwidth]{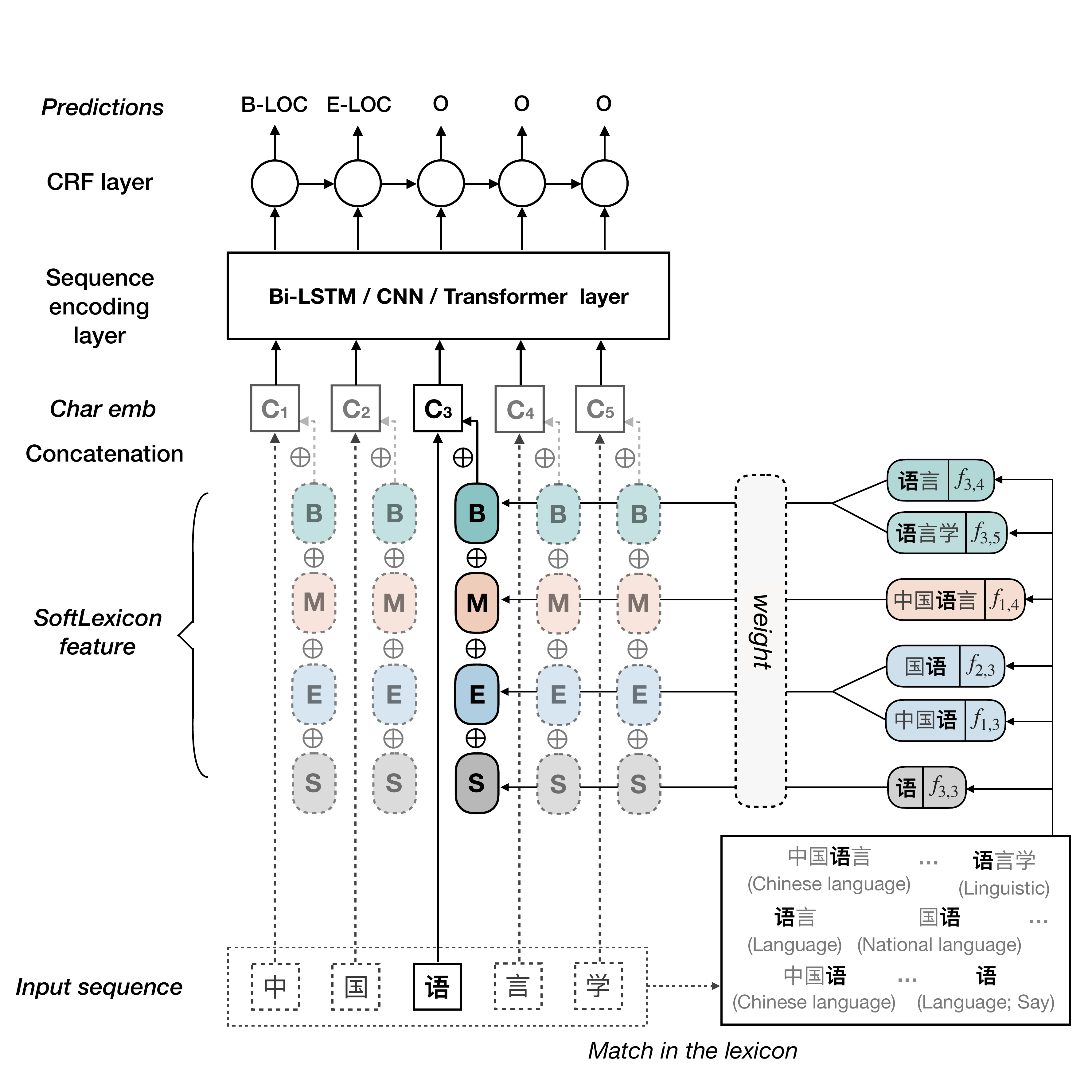}
\caption{The overall architecture of the proposed method.}
\label{fig:model_figure}
\end{figure}

In this work, we sought to retain the merits of Lattice-LSTM while overcoming its drawbacks.
% , i.e., we simplify the model architecture to achieve a faster inference speed. 
To this end, we propose a novel method in which lexicon information is introduced by simply adjusting the character representation layer of an NER model. We refer to this method as \textit{SoftLexicon}. As shown in Figure \ref{fig:model_figure}, the overall architecture of the proposed method is as follows. First, each character of the input sequence is mapped into a dense vector. Next, the SoftLexicon feature is constructed and added to the representation of each character.
% , consisting of three steps, including mapping and categorizing the matched words, condensing the word sets and combining with the character representation. 
Then, these augmented character representations are put into the sequence modeling layer and the CRF layer to obtain the final predictions.
% As shown in Figure \ref{fig:model_figure}, the overall architecture of our method consists of four parts. First, we map each character of the input sequence into a dense vector. Second, we introduce the lexicon information into the character representation with the proposed 
% We denote the input sequence as $\s=\{c_1, c_2, ..., c_n\}$, and denote the sub-sequence $\{c_i, c_{i+1}, ..., c_j\}$ of $\bm{S}$ as $w_{i,j}$.

\subsection{Character Representation Layer}

For a character-based Chinese NER model, the input sentence is seen as a character sequence $s=\{c_1, c_2, \cdots, c_n\} \in \mathcal{V}_c$, where $\mathcal{V}_c$ is the character vocabulary. Each character $c_i$ is represented using a dense vector (embedding):
\begin{equation}
    \bm{x}_i^c = \bm{e}^{c} (c_i),
\end{equation}
where $\bm{e}^{c}$ denotes the character embedding lookup table. 

\paragraph{Char + bichar.} In addition, \citeauthor{zhang2018chinese}, \shortcite{zhang2018chinese} has proved that character bigrams are useful for representing characters, especially for those methods {not} using word information. Therefore, it is common to augment the character representations with bigram embeddings:
\begin{equation}
    \bm{x}_i^c = [\bm{e}^{c} (c_i) ; \bm{e}^{b} (c_i, c_{i+1})],
\end{equation}
where $\bm{e}^{b}$ denotes the bigram embedding lookup table. %The sequence of character representations $\bm{x}_i^c$ form the matrix representation $\bm{x}^s=\{\bm{x}_1^c, \cdots, \bm{x}_n^c\}$ of $s$. 

\subsection{Incorporating Lexicon Information} 
The problem with the purely character-based NER model is that it fails to exploit word information. To address this issue, we proposed two methods, as described below, to introduce the word information into the character representations. In the following, for any input sequence $s=\{c_1, c_2, \cdots, c_n\}$, $w_{i,j}$ denotes its sub-sequence $\{c_i, c_{i+1}, \cdots, c_j\}$.

\subsubsection{ExSoftword Feature}

The first conducted method is an intuitive extension of the Softword method, called \textit{ExSoftword}. Instead of choosing one segmentation result for each character, it proposes to retain all possible segmentation results obtained using the lexicon: 
% talk about the general method
\begin{equation}
\bm{x}^c_j \leftarrow [\bm{x}^c_j; \bm{e}^{seg}(segs(c_j)],
\end{equation}
where $segs(c_j)$ denotes all segmentation labels related to $c_j$, and $\bm{e}^{seg}(segs(c_j))$ is a 5-dimensional multi-hot vector with each dimension corresponding to an item of $\{\text{B, M, E, S, O\}}$. 

\begin{figure}
\centering
\includegraphics[width=1\columnwidth]{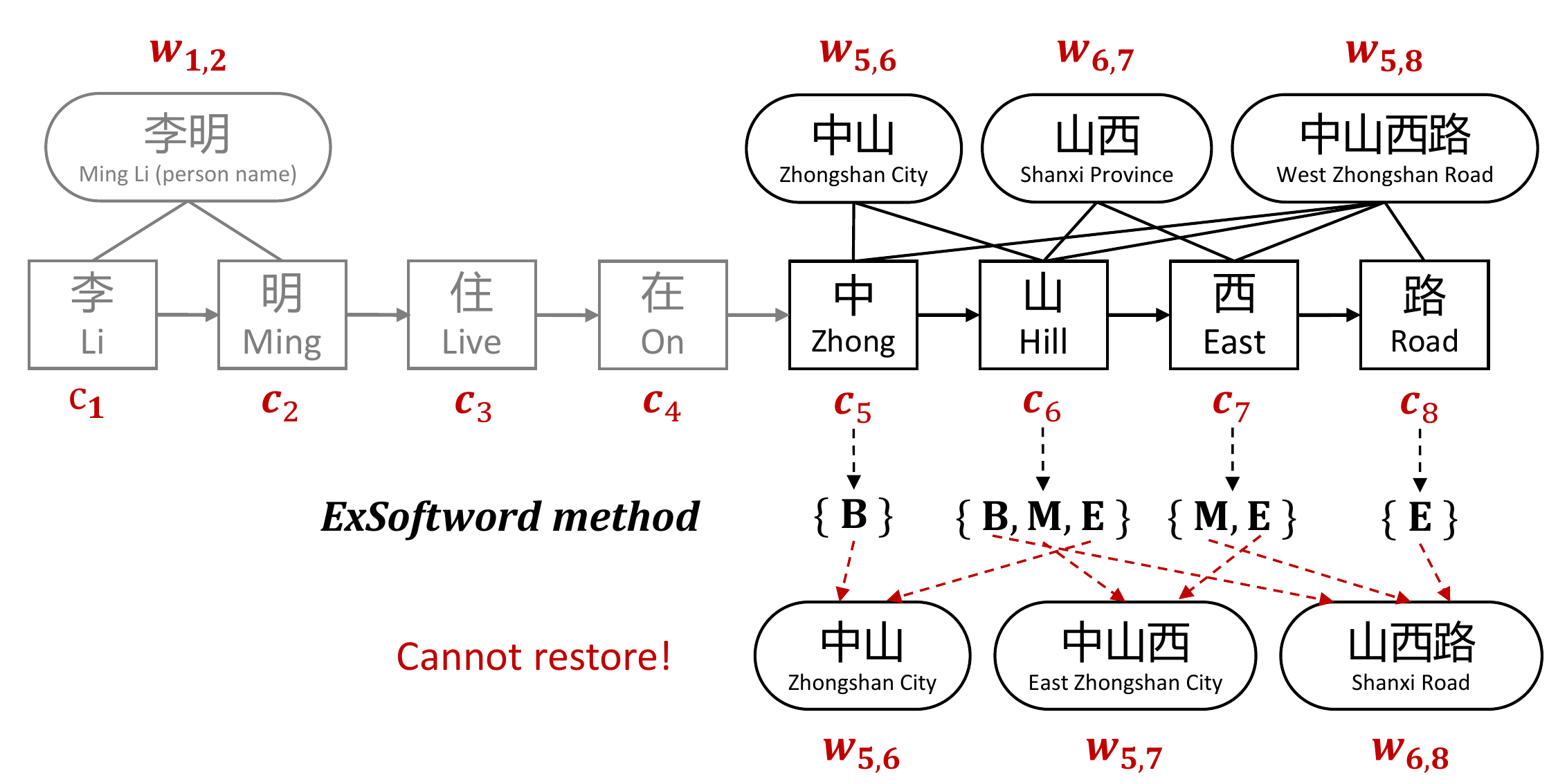}
\caption{The ExSoftword method. }
\label{fig:exsoft_figure}
\end{figure}

As an example presented in Figure \ref{fig:exsoft_figure}, the character $c_7$ (``西") occurs in two words, $w_{5,8}$ (``中山西路") and $w_{6,7}$ (``山西"), that match the lexicon, and it occurs in the middle of ``中山西路" and the end of ``山西". Therefore, its corresponding segmentation result is $\{\text{M}, \text{E}\}$, and
% Different from the Softword strategy, we use a lexicon to segment the input sentence, and preserve all segmentation results. As an example presented in Figure \ref{fig:exsoft_figure}, given the input sequence and the lexicon matching results, the Exsoftword features for character $c_7$ is $\{\text{M}, \text{E}\}$ since there exists lexicon matched words $w_{5,8}$ (``中山西路") that contains it inside, as well as $w_{6,7}$ (``山西"), which ends with it.
its character representation is enriched as follows:
\begin{equation}
\bm{x}^c_7 \leftarrow [\bm{x}^c_7; \bm{e}^{seg}(\{M, E\})].
\end{equation}
Here, the second and third dimensions of $\bm{e}^{seg}(\cdot)$ are set to 1, and the rest dimensions are set to 0.

The problem of this approach is that it cannot fully inherit the two merits of Lattice-LSTM. First, it fails to introduce pre-trained word embeddings. Second, it still losses information of the matching results.
As shown in Figure \ref{fig:exsoft_figure}, the constructed ExSoftword feature for characters $\{c_{5}, c_{6}, c_{7}, c_{8}\}$ is $\{\{\text{B}\}, \{\text{B},\text{M},\text{E}\}, \{\text{M},\text{E}\},\{\text{E}\}\}$. However, given this constructed sequence, there exists more than one corresponding matching results, such as \{$w_{5,6}$ (``中山"), $w_{5,7}$ (``中山西"), $w_{6,8}$ (``山西路")\} and \{$w_{5,6}$ (``中山"), $w_{6,7}$ (``山西"), $w_{5,8}$ (``中山西路")\}. Therefore, we cannot tell which is the correct result to be restored.
% ``中山", ``山西" and ``中山西路". 
% there exists more than one corresponding matching results, such that the initial matching result cannot be restored.

\subsubsection{SoftLexicon} 
\label{sec:method}
Based on the analysis on Exsoftword, we further developed the \textit{SoftLexicon} method to incorporate the lexicon information.
% preserve not only the possible segmentation labels of each character but also the corresponding matched words. 
%We refer to this method as \textit{Soft-lexicon}. 
The SoftLexicon features are constructed in three steps. 

\begin{figure}
\centering
\includegraphics[width=1\columnwidth]{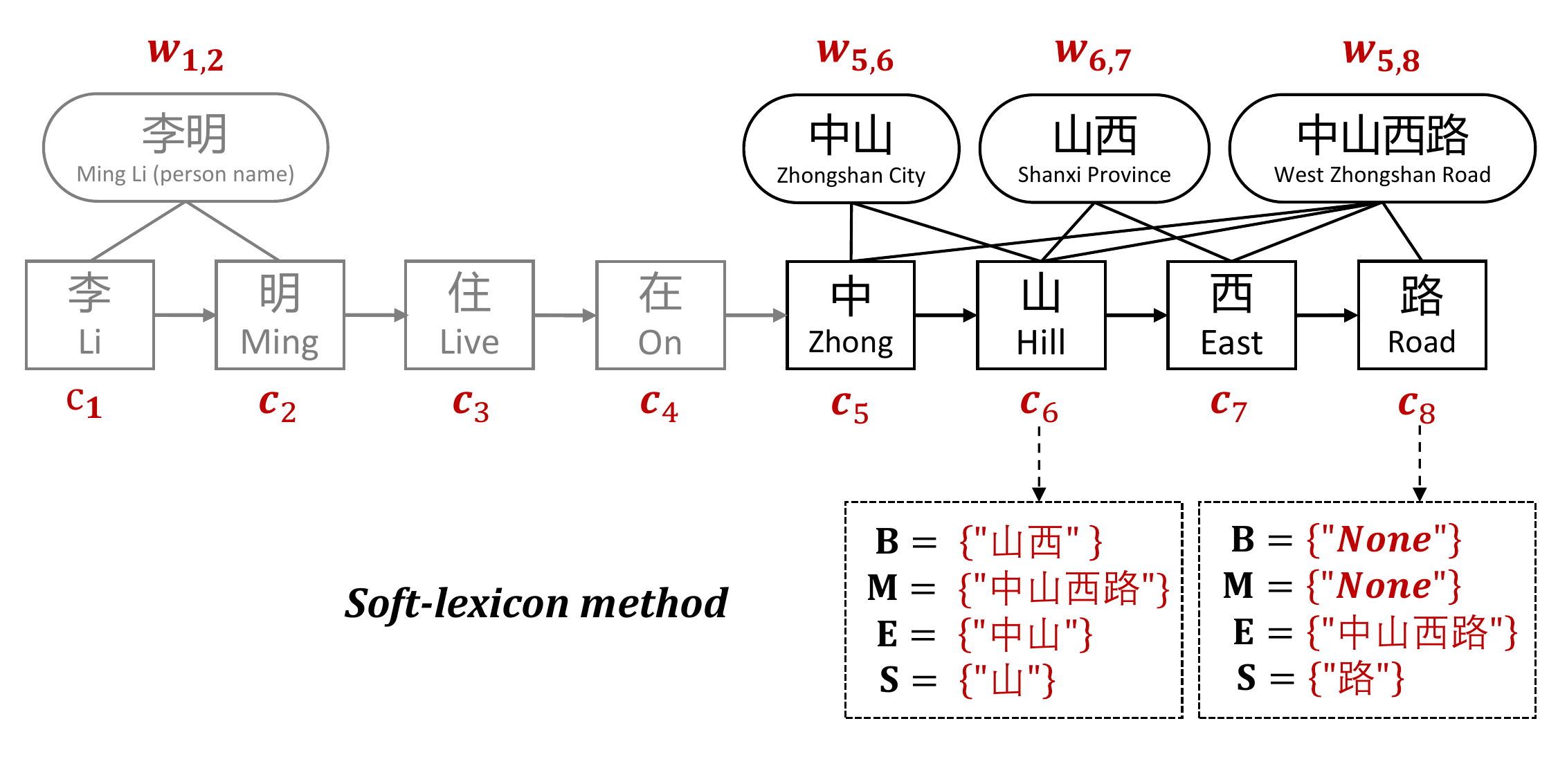}
\caption{The SoftLexicon method. }
\label{fig:softlexicon_figure}
\end{figure}

\paragraph{Categorizing the matched words.}
First, to retain the segmentation information, all matched words of each character $c_i$ is categorized into four word sets ``BMES", which is marked by the four segmentation labels. For each character $c_i$ in the input sequence $s=\{c_1, c_2, \cdots, c_n\}$, the four set is constructed by:
\begin{equation}\label{eq:set_definition}
\begin{split}
&\rm{B}(c_i) = \{w_{i,k}, \forall w_{i,k} \in \rm{L}, i < k \leq n\}, \\
&\rm{M}(c_i) = \{w_{j,k}, \forall w_{j,k} \in \rm{L}, 1 \leq j < i < k \leq n\}, \\
&\rm{E}(c_i) = \{w_{j,i}, \forall w_{j,i} \in \rm{L}, 1 \leq j < i \}, \\
&\rm{S}(c_i) = \{c_i, \exists c_i \in \rm{L} \}.
\end{split}
\end{equation}
Here, $\rm{L}$ denotes the lexicon we use in this work.
% , and $w_{i,j}$ is a sub-sequence of the input sequence. 
% For example, the word set $\rm{B}(c_i)$ consists of all sub-sequences of $s$ that match the lexicon and begin with $c_i$, and $\rm{S}(c_i)$ is a single-character word comprising of $c_i$.
Additionally, if a word set is empty, a special word ``NONE" is added to the empty word set. An example of this categorization approach is shown in Figure \ref{fig:softlexicon_figure}. Noted that in this way, not only we can introduce the word embedding, but also no information loss exists since the matching results can be exactly restored from the four word sets of the characters.

\paragraph{Condensing the word sets.}
After obtaining the ``BMES" word sets for each character, each word set is then condensed into a fixed-dimensional vector. In this work, we explored two approaches for implementing this condensation.

The \textbf{first} implementation is the intuitive mean-pooling method:
\begin{equation}\label{eq:meanpool}
\bm{v}^{s}(\mathcal{S}) = \frac{1}{|\mathcal{S}|}\sum_{w \in \mathcal{S}} \bm{e}^w(w).
\end{equation}
Here, $\mathcal{S}$ denotes a word set and $\bm{e}^w$ denotes the word embedding lookup table. 

However, as shown in Table \ref{tab:ablation}, the results of empirical studies revealed that this algorithm does not perform well. Therefore, a weighting algorithm is introduced to further leverage the word information. To maintain computational efficiency, we did not opt for a dynamic weighting algorithm like attention. 
% Therefore, it propose to weigh the embedding of each word in the word set to obtain the pooling representation. However, considering computational efficiency, the dynamic weighting algorithm, like attention, to obtain the weight of each word is not chosen. 
Instead, we propose using the frequency of each word as an indication of its weight. 
% The basic rationale for this is that the more times a sub-sequence occurs in the data, the more likely it is a word. 
Since the frequency of a word is a static value that can be obtained offline, this can greatly accelerate the calculation of the weight of each word. 

Specifically, let $z(w)$ denote the frequency that a lexicon word $w$ occurs in the statistical data, the weighted representation of the word set $\mathcal{S}$ is obtained as follows:
\begin{equation}\label{eq:weight_meanpool}
\bm{v}^s(S) = \frac{4}{Z} \sum_{w \in S} z(w) \bm{e}^w(w),
\end{equation}
where $$Z = \sum_{w \in \rm{B} \cup \rm{M} \cup \rm{E} \cup \rm{S}} z(w).$$ Here, weight normalization is performed on all words in the four word sets to make an overall comparison.
% across the four word sets.

In this work, the statistical data set is constructed from a combination of training and developing data of the task. Of course, if there is unlabelled data in the task, the unlabeled data set can serve as the statistical data set. In addition, note that the frequency of $w$ does not increase if $w$ is covered by another sub-sequence that matches the lexicon. This prevents the problem in which the frequency of a shorter word is always less than the frequency of the longer word that covers it.

\paragraph{Combining with character representation.}
The final step is to combine the representations of four word sets into one fix-dimensional feature, and add it to the representation of each character. In order to retain as much information as possible, we choose to concatenate the representations of the four word sets, and the final representation of each character is obtained by:
\begin{equation}\label{eq:combine_rep}
\begin{split}
&\resizebox{0.85\columnwidth}{!}{$\bm{e}^{s}\left(\rm{B}, \rm{M}, \rm{E}, \rm{S}\right) =  \left[\bm{v}^{s}(\mathrm{B}) ; \bm{v}^{s}(\rm{M})  ; \bm{{v}}^s(\rm{E})  ; \bm{{v}}^s(\rm{S})\right]$}, \\
& \quad \quad \quad \quad \bm{x}^c \leftarrow [\bm{x}^c; \bm{e}^{s}(\rm{B}, \rm{M}, \rm{E}, \rm{S})].
\end{split}
\end{equation}
Here, $\bm{v}^s$ denotes the weighting function above.

\subsection{Sequence Modeling Layer} \label{sec:sequence_modeling}
With the lexicon information incorporated, the character representations are then put into the sequence modeling layer, which models the dependency between characters. Generic architectures for this layer including the bidirectional long-short term memory network(BiLSTM), the Convolutional Neural Network(CNN) and the transformer\cite{vaswani2017attention}. In this work, we implemented this layer with a single-layer Bi-LSTM.

Here, we precisely show the definition of the forward LSTM:
\begin{equation}
\begin{split}
\begin{bmatrix}
\bm{i}_t \\ \bm{f}_t \\ \bm{o}_t \\ \bm{\widetilde{c}}_t \\
\end{bmatrix} &= \begin{bmatrix}
\sigma \\ \sigma \\ \sigma \\ \tanh \\
\end{bmatrix} \left( \bm{W} \begin{bmatrix}
\bm{x}_t^c \\ \bm{h}_{t-1} \\
\end{bmatrix} + \bm{b} \right), \\
\bm{c}_t &= \bm{\widetilde{c}}_t \odot \bm{{i}}_t + \bm{c}_{t-1} \odot \bm{{f}}_t,\\
\bm{h}_t &= \bm{o}_t \odot \tanh(\bm{c_t}).\\
\end{split}
\end{equation}
where $\sigma$ is the element-wise sigmoid function and $\odot$ represents element-wise product. $\bm{\mathrm{W}}$ and $\bm{\mathrm{b}}$ are trainable parameters. The backward LSTM shares the same definition as the forward LSTM yet model the sequence in a reverse order. 
The concatenated hidden states at the $i^{th}$ step of the forward and backward LSTMs $\bm{h}_i=[\overrightarrow{\bm{h}}_i ; \overleftarrow{\bm{h}}_i]$ forms the context-dependent representation of $c_i$.

\subsection{Label Inference Layer}
On top of the sequence modeling layer, it is typical to apply a sequential conditional random field (CRF) \cite{lafferty2001conditional} layer to perform label inference for the whole character sequence at once:
\begin{equation}
p(\bm{y}|{s}; \bm{\theta}) = \frac{\prod_{t=1}^n \phi_{t}(\bm{y}_{t-1}, \bm{y}_t|{s})}{\sum_{\bm{y}^\prime \in \mathcal{Y}_s} \prod_{t=1}^n \phi_{t}(\bm{y}^{\prime}_{t-1}, \bm{y}^{\prime}_t|{s})}.
\end{equation}
Here, $\mathcal{Y}_s$ denotes all possible label sequences of $s$, and $\phi_{t}({y}^\prime, {y}|{s})=\exp(\bm{w}^T_{{y}^\prime, {y}} \bm{h}_t + b_{{y}^\prime, {y}})$, where $\bm{w}_{{y}^\prime, {y}}$ and $ b_{{y}^\prime, {y}}$ are trainable parameters corresponding to the label pair $({y}^\prime, {y})$, and $\bm{\theta}$ denotes model parameters. 
For label inference, it searches for the label sequence $\bm{y}^{*}$ with the highest conditional probability given the input sequence ${s}$:
\begin{equation}
\bm{y}^{*} = \mathop{\arg\max}_{\bm{y}} p(\bm{y}|{s}; \bm{\theta}),
\end{equation}
which can be efficiently solved using the Viterbi algorithm \cite{forney1973viterbi}.

\begin{table}[t]
\centering
\resizebox{\columnwidth}{!}{
\begin{tabular}{lcccc}
  \textbf{Datasets} & \textbf{Type} & \textbf{Train} & \textbf{Dev} & \textbf{Test} \\
  \hline
  \hline
  \multirow{2}{*}{OntoNotes} & Sentence  & 15.7k & 4.3k & 4.3k \\
  & Char & 491.9k & 200.5k & 208.1k \\
  \hline
  \multirow{2}{*}{MSRA} & Sentence  & 46.4k & - & 4.4k \\
  & Char & 2169.9k & - & 172.6k \\
  \hline
  \multirow{2}{*}{Weibo} & Sentence  & 1.4k & 0.27k & 0.27k \\
  & Char & 73.8k & 14.5 & 14.8k \\
  \hline
  \multirow{2}{*}{Resume} & Sentence  & 3.8k & 0.46 & 0.48k \\
  & Char & 124.1k & 13.9k & 15.1k \\
  \hline
\end{tabular}
}
\caption{Statistics of datasets.}
  \label{tab:data_statistic}
\end{table}

\section{Experiments} \label{sec:experiment}
%Through the experiments, we evaluate the effectiveness of our method in performance and inference speed primary compared with Lattice-LSTM. In addition, we verify the applicability of our method to different sequence modeling architectures.

% \subsection{Experiment Design}
% \textbf{First}, we performed a development study on our method with the LSTM-based sequence modeling layer to compare the implementations of $\bm{v}^s$. Decisions made in this step will be applied to the following experiments. 
% \textbf{Second}, we verified the computational efficiency of our method compared with Lattice-LSTM and LR-CNN \cite{guicnn}, which is a follower of Lattice-LSTM for faster inference speed.
% \textbf{Third}, we verified the effectiveness of our method by comparing its performance with that of Lattice-LSTM and other comparable models on four Chinese NER benchmarks. \textbf{Lastly}, we verified the applicability of our method to different sequence model architectures.

\subsection{Experiment Setup}
Most experimental settings in this work followed the protocols of Lattice-LSTM \cite{zhang2018chinese}, including tested datasets, compared baselines, evaluation metrics (P, R, F1), and so on. To make this work self-completed, we concisely illustrate some primary settings of this work. 
% \footnote{The source code of this paper will be released at https://github.com/xxx.}

\begin{table}[h!]
    \centering
    \resizebox{\columnwidth}{!}{
    \begin{tabular}{lcccc}
          \textbf{Models} & \textbf{OntoNotes} & 
          \textbf{MSRA} & 
          \textbf{Weibo} & 
          \textbf{Resume}\\  \hline \hline
          Lattice-LSTM & 1$\times$ & 1$\times$ & 1$\times$ & 1$\times$ \\
          LR-CNN (\citeauthor{gui2019cnn}, 2019) & 2.23$\times$ &  1.57$\times$ & 2.41$\times$ & 1.44$\times$ \\
          BERT-tagger  & 2.56$\times$ & 2.55$\times$ & 4.45$\times$ & 3.12$\times$ \\ 
          BERT + LSTM + CRF  & 2.77$\times$ & 2.32$\times$ & 2.84$\times$ & 2.38$\times$ \\ 
          \hline
          SoftLexicon (LSTM)  &6.15$\times$ & 5.78$\times$ & 6.10$\times$ & 6.13$\times$ \\
          SoftLexicon (LSTM) + bichar  & 6.08$\times$ & 5.95$\times$ & 5.91$\times$ & 6.45$\times$ \\ 
          SoftLexicon (LSTM) + BERT & 2.74$\times$ & 2.33$\times$ & 2.85$\times$ & 2.32$\times$ \\ 
        %   \hline
        %   Soft-lexicon (LSTM)  &6.15$\times$ & 5.78$\times$ & 6.10$\times$ & 6.13$\times$ \\
        %   Soft-lexicon(LSTM) + bichar  & 6.08$\times$ & 5.95$\times$ & 5.91$\times$ & 6.45$\times$ \\         
        %   171.46 & 210.84 & 143.45 & 215.27
          \hline
    \end{tabular}}
    \caption{Inference speed (average sentences per second, the larger the better) of our method with LSTM layer compared with Lattice-LSTM, LR-CNN and BERT.}
    \label{tab:speed}
%\vspace*{-0.5\baselineskip}
\end{table}

\subsubsection{Datasets}
The methods were evaluated on four Chinese NER datasets, including OntoNotes~\cite{weischedel2011ontonotes}, MSRA~\cite{levow2006third}, Weibo NER~\cite{peng2015named,he2017f}, and Resume NER~\cite{zhang2018chinese}. OntoNotes and MSRA are from the newswire domain, where gold-standard segmentation is available for training data. For OntoNotes, gold segmentation is also available for development and testing data. Weibo NER and Resume NER are from social media and resume, respectively. There is no gold standard segmentation in these two datasets. 
Table \ref{tab:data_statistic} shows statistic information of these datasets.
As for the lexicon, we used the same one as Lattice-LSTM, which contains 5.7k single-character words, 291.5k two-character words, 278.1k three-character words, and 129.1k other words. In addition, the pre-trained character embeddings we used are also the same with Lattice-LSTM, which are pre-trained on Chinese Giga-Word using word2vec.

\subsubsection{Implementation Detail}

In this work, we implement the sequence-labeling layer with Bi-LSTM. Most implementation details followed those of Lattice-LSTM, including character and word embedding sizes, dropout, embedding initialization, and LSTM layer number. Additionally, the hidden size was set to 200 for small datasets Weibo and Resume, and 300 for larger datasets OntoNotes and MSRA. The initial learning rate was set to 0.005 for Weibo and 0.0015 for the rest three datasets with Adamax \cite{kingma2014adam} step rule
% When applying the CNN- and transformer- based sequence modeling layers, most hyper-parameters were the same as those used in the LSTM-based model. In addition, the layer number $L$ for the CNN-based model was set to 4, and that for  transformer-based model was set to 2 with h=4 parallel attention layers. Kernel number $k_f$ of the CNN-based model was set to 512 for MSRA and 128 for the other datasets in all layers
\footnote{Please refer to the attached source code for more implementation detail of this work and access \url{https://github.com/jiesutd/LatticeLSTM} for pre-trained word and character embeddings.}.

% Figure \ref{fig:study_bichar} shows the F1 score of our method against the number of training iterations, with and without using character bigrams. From the figure, we can see that the introduction of character bigrams does not significantly improve our method. A possible explanation for this is that the word information introduced by our proposed method has covered the bichar information. \textit{Therefore, in the following experiments, we did not use bichar in our method.}

% \begin{figure}
% \centering
% \includegraphics[width=1\columnwidth]{char_vs_bichar.pdf}
% \caption{F1 of our proposed method against the number of training iterations on OntoNotes when using bichar or not.}
% \label{fig:study_bichar}
% \end{figure}

\subsection{Computational Efficiency Study}

Table \ref{tab:speed} shows the inference speed of the SoftLexicon method when implementing the sequence modeling layer with a bi-LSTM layer. The speed was evaluated based on the average number of sentences processed by the model per second using a GPU (NVIDIA TITAN X). 
% For a fair comparison with Lattice-LSTM and LR-CNN, \textit{we set the batch size of our method to 1 at inference time}. %From the table, we can see that the inference speed of our method was much faster than that of Lattice-LSTM when using the LSTM-based sequence modeling layer, and it was also much faster than that of LR-CNN, which used a CNN-based architecture to implement the sequence modeling layer. And as expected, the inference speed of our method with the CNN-based sequence modeling layer was somehow faster than that of our method with the LSTM-based and Transformer-based sequence model layer. 
% From the table, we can first observe that when implementing the sequence modeling layer 
% with the CNN-based architecture, our method achieved the fastest inference speed.
% However, even with the slower LSTM-based architecture, our method achieved a much 
% faster inference speed than Lattice-LSTM and LR-CNN, which used a CNN-based architecture
% to implement the sequence modeling layer. These results illustrated the advantage of our method with respect to inference speed. 
From the table, we can observe that when decoding with the same batch size (=1), the proposed method is considerably more efficient than Lattice-LSTM and LR-CNN, performing up to 6.15 times faster than Lattice-LSTM. The inference speeds of SoftLexicon(LSTM) with bichar are close to those without bichar, since we only concatenate an additional feature to the character representation. The inference speeds of the BERT-Tagger and SoftLexicon (LSTM) + BERT models are limited due to the deep layers of the BERT structure. However, the speeds of the SoftLexicon (LSTM) + BERT model are still faster than those of Lattice-LSTM and LR-CNN on all datasets.
% Moreover, the proposed model does not suffer from the disability of parallel training, which is an unignorable drawback of Lattice-LSTM for practical usage. The results show that when the inference batch size is set to 4, the proposed method achieves up to 14.30 times the speed of Lattice-LSTM. 
% These results illustrated the advantage of our method with respect to inference speed. 

\begin{figure}
\centering
% \begin{minipage}{3.8cm}
% \centering
% \includegraphics[width=1\columnwidth]{length_f1.pdf}
% \caption{World Map}
% \end{minipage}
% \begin{minipage}{3.8cm}
% \centering
\includegraphics[width=0.9\columnwidth]{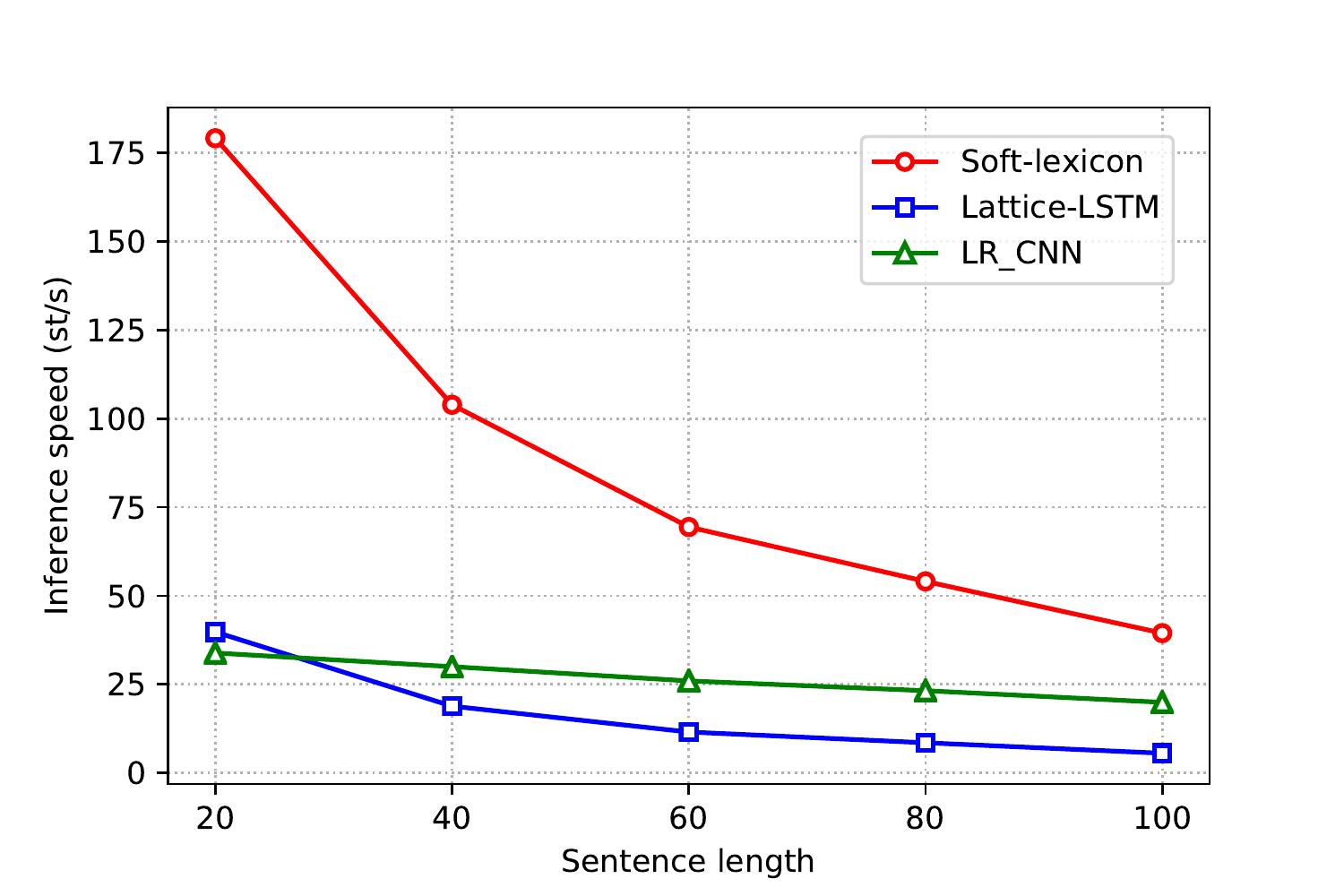}
% \caption{Concrete and Constructions}
% \end{minipage}
\caption{Inference speed against sentence length. We use a same batch size of 1 for a fair speed comparison. }
\label{fig:lengthtest}
\end{figure}

To further illustrate the efficiency of the SoftLexicon method, we also conducted an experiment to evaluate its inference speed against sentences of different lengths, as shown in Table \ref{fig:lengthtest}. For a fair comparison, we set the batch size to 1 in all of the compared methods. The results show that the proposed method achieves significant improvement in speed over Lattice-LSTM and LR-CNN when processing short sentences. With the increase of sentence length, the proposed method is consistently faster than Lattice-LSTM and LR-CNN despite the speed degradation due to the recurrent architecture of LSTM. Overall, the proposed SoftLexicon method shows a great advantage over other methods in computational efficiency.

% \subsection{Development Experiments}
% \label{development_experiments}
% In this experiment, we compared the performance of our Soft-lexicon method with different implementations of $\bm{v}^s$, when the sequence model layer was implemented with bidirectional LSTM. 
% Table \ref{tab:imp_result} shows the performance of our method when implementing $\bm{v}^s$ with the mean-pooling algorithm and the weighted pooling algorithm without using character bigrams. From the table, we can see that the weighted pooling algorithm performed much better than the mean-pooling algorithm, verifying the effectiveness of our word weighting strategy.
% \textit{Therefore, in the following experiments, we in default applied the weighted pooling algorithm to implement $\bm{v}^s$}.

% \begin{table}[t]
% \centering
% \resizebox{.6\columnwidth}{!}{
% \begin{tabular}{|l|c|c|c|}
% \hline
%     Dataset     & MP     & WP       \\ \hline 
%     NoteNotes   & 0.7257 & \textbf{0.7564} \\
%     MSRA        & 0.9276 & \textbf{0.9366} \\
%     Weibo       & 0.5772 & \textbf{0.6142}\\
%     Resume      & 0.9533 & \textbf{0.9553}\\ \hline \hline
%     Average     & 0.7560 & \textbf{0.8157} \\
%     \hline
% \end{tabular}
% }
% \caption{F1 score of our method with different implementations of $\bm{v}^s$. MP denotes the mean-pooling algorithm defined in Eq. (\ref{eq:meanpool}), WP denotes the frequency weighted pooling algorithm defined in Eq. (\ref{eq:weight_meanpool})}
% \label{tab:imp_result}
% %\vspace*{-\baselineskip}
% \end{table}

\subsection{Effectiveness Study}

Tables \ref{tab:ontonote}$-$\ref{tab:resume}\footnote{In Table  \ref{tab:ontonote}$-$\ref{tab:weibo}, $*$ indicates that the model uses external labeled data for semi-supervised learning. 
$\dagger$ means that the model also uses discrete features.} show the performances of our method against the compared baselines. In this study, the sequence modeling layer of our method was implemented with a single layer bidirectional LSTM.

\begin{table}[t]
\centering
\resizebox{\columnwidth}{!}{
\begin{tabular}{|l|l|c|c|c|}
  \hline
  \textbf{Input}  & \textbf{Models} & \textbf{P} & \textbf{R} & \textbf{F1}\\
  \hline
  \multirow{6}{*}{Gold seg} & \citeauthor{yang2016combining}, \citeyear{yang2016combining} & 65.59  & 71.84 & 68.57 \\
  & \citeauthor{yang2016combining}, \citeyear{yang2016combining}$^{*\dagger}$ & 72.98 & \textbf{80.15} & \textbf{76.40} \\
  & \citeauthor{che2013named}, \citeyear{che2013named}$^{*}$ & {77.71} & 72.51 & 75.02 \\
  & \citeauthor{wang2013effective}, \citeyear{wang2013effective}$^{*}$  & 76.43 & 72.32 & 74.32 \\
  \cline{2-5}
  & Word-based (LSTM) & 76.66 & 63.60 & 69.52 \\
  & \ \ + char + bichar & \textbf{78.62} & {73.13} & {75.77} \\
  \hline
  \hline
  \multirow{2}{*}{Auto seg} & Word-based (LSTM) & 72.84 & 59.72 & 65.63 \\
  & \ \ + char + bichar & 73.36 & {70.12} & {71.70} \\
  \hline
  \multirow{4}{*}{No seg} & Char-based (LSTM) & 68.79 & 60.35 & 64.30 \\
  %& \ \ + bichar & 72.31 & 66.47  & 69.27  \\
  & \ \ + bichar + softword & 74.36 & 69.43 & 71.89 \\
  & \ \ + ExSoftword & 69.90 & 66.46 & 68.13\\
  & \ \ + bichar + ExSoftword & 73.80 & 71.05 & 72.40\\
  & Lattice-LSTM & 76.35 & 71.56 & 73.88 \\ 
   & LR-CNN (\citeauthor{gui2019cnn}, 2019) & 76.40 & 72.60 & 74.45 \\ 
%   & Proposed (LSTM) & \textbf{77.94} & \textbf{72.46} & \textbf{75.10} \\
%   & Proposed (LSTM) & {77.31} & {73.85} & {75.54} \\
  & SoftLexicon (LSTM) & {77.28} & {74.07} & {75.64} \\
    & SoftLexicon (LSTM) + bichar & {77.13} & \textbf{75.22} & \textbf{76.16} \\
  \cline{2-5}
%   & BERT-Tagger & 78.01 & 80.35 & 79.16 \\
& BERT-Tagger & 76.01 & 79.96 & 77.93 \\
   & BERT + LSTM + CRF & 81.99 & 81.65 & 81.82 \\
%   & Lattice-Transformer + BERT & 79.62 & 81.82 & 80.60 \\
  & SoftLexicon (LSTM) + BERT & \textbf{83.41} & \textbf{82.21} & \textbf{82.81} \\
  \hline
\end{tabular}}
\caption{Performance on OntoNotes. A model followed by (LSTM) (e.g., Proposed (LSTM)) indicates that its sequence modeling layer is LSTM-based.}
  \label{tab:ontonote}
\end{table}

\paragraph{OntoNotes.} Table \ref{tab:ontonote} shows results \footnote{A result in boldface indicates that it is statistically significantly better ($p<0.01$ in pairwise $t-$test) than the others in the same box.} on the OntoNotes dataset, where gold word segmentation is provided for both training and testing data. 
%The methods of the ``Gold seg" and "Auto seg" group are word-based that build on the gold word segmentation results and the automatic segmentation results, respectively. The automatic segmentation results were generated by the segmenter trained on training data of OntoNotes. Methods of the "No seg" group are character-based.
The methods of the ``Gold seg" and the ``Auto seg" groups are all word-based, with the former input building on gold word segmentation results and the latter building on automatic word segmentation results by a segmenter trained on OntoNotes training data. The methods used in the ``No seg" group are character-based.
From the table, we can make several observations. \textbf{{First}}, when gold word segmentation was replaced by automatically generated word segmentation, the F1 score decreases from 75.77\% to 71.70\%. This reveals the problem of treating the predicted word segmentation result as the true result in the word-based Chinese NER. \textbf{{Second}}, the F1 score of the Char-based (LSTM)+ExSoftword model is greatly improved from that of the Char-based (LSTM) model. This indicates the feasibility of the naive ExSoftword method. However, it still greatly underperforms relative to Lattice-LSTM, which reveals its deficiency in utilizing word information. \textbf{{Lastly}}, the proposed SoftLexicon method outperforms Lattice-LSTM by 1.76\% with respect to the F1 score, and obtains a greater improvement of 2.28\% combining the bichar feature. It even performs comparably with the word-based methods of the ``Gold seg" group, verifying its effectiveness on OntoNotes.

%the dataset (Che et al., 2013; Wang et al., 2013), which leverage bilingual data.  

\begin{table}[t]
    \centering
    \resizebox{.95\columnwidth}{!}{
    \begin{tabular}{lccc}
          \textbf{Models} & \textbf{P} & \textbf{R} & \textbf{F1} \\ \hline \hline
            \citeauthor{chen2006chinese}, \citeyear{chen2006chinese} & 91.22  & 81.71 & 86.20 \\
            \citeauthor{zhang2006word} \citeyear{zhang2006word}$^*$ & 92.20 & 90.18 & 91.18 \\
            \citeauthor{zhou2013chinese} \citeyear{zhou2013chinese} & 91.86 & 88.75 & 90.28 \\
            \citeauthor{lu2016multi} \citeyear{lu2016multi}  & - & - & 87.94 \\
            \citeauthor{dong2016character} \citeyear{dong2016character} & 91.28 & 90.62 & 90.95 \\ \hline   
        %   Word-based (LSTM) & 90.57 & 83.06 & 86.65 \\
        %   \ \ +char+bichar  & 91.05 & 89.53 & 90.28 \\
          Char-based (LSTM) & 90.74 & 86.96 & 88.81 \\
          \ \ + bichar+softword & 92.97 & 90.80 &91.87 \\ 
          \ \ + ExSoftword & 90.77 & 87.23 & 88.97\\
          \ \ + bichar+ExSoftword & 93.21 & 91.57 & 92.38\\
          Lattice-LSTM & 93.57 & 92.79 & 93.18 \\ 
          LR-CNN (\citeauthor{gui2019cnn}, 2019) & {94.50} & {92.93} & {93.71} \\ 
        %   proposed (LSTM) & 93.56 & \textbf{93.44} & {93.50}
          SoftLexicon (LSTM) & 94.63 & {92.70} & {93.66}\\
          SoftLexicon (LSTM) + bichar & \textbf{94.73} & \textbf{93.40} & \textbf{94.06}\\ \hline
          BERT-Tagger & 93.40 & {94.12} & 93.76 \\  % 94.56 & 95.53 & 95.04
          BERT + LSTM + CRF & 95.06 & 94.61 & 94.83 \\
%  Lattice-Transformer + BERT & - & - & 94.25 \\
  SoftLexicon (LSTM) + BERT & \textbf{95.75} & \textbf{95.10} & \textbf{95.42} \\ \hline
          
    \end{tabular}}
    \caption{Performance on MSRA.}
    \label{tab:msra}
\end{table}

% \paragraph{MSRA.} Table \ref{tab:msra} shows results on MSRA. The word-based methods were built on the automatic segmentation results generated by the segmenter trained on training data of MSRA.
% Compared methods includes the best statistical models on this data set, which leveraged rich handcrafted features \cite{chen2006chinese,zhang2006word,zhou2013chinese}, character embedding features \cite{lu2016multi}, and radical features \cite{dong2016character}. From the table, we can observe that our method also outperforms Lattice-LSTM and performs comparatively with LR-CNN. This verifies the effectiveness of our method on MSRA. 

\begin{table}[t]
    \centering
    \resizebox{0.95\columnwidth}{!}{
    \begin{tabular}{lcccc}
          \textbf{Models} & \textbf{NE} & \textbf{NM} & \textbf{Overall} \\ \hline \hline
            \citeauthor{peng2015named}, \citeyear{peng2015named} & 51.96  & 61.05 & 56.05 \\
            \citeauthor{peng2016improving}, \citeyear{peng2016improving}$^*$ & \textbf{55.28} & \textbf{62.97} & \textbf{58.99} \\
            \citeauthor{he2017f}, \citeyear{he2017f} & 50.60 & 59.32 & 54.82 \\
            \citeauthor{he2017unified}, \citeyear{he2017unified}$^*$  & 54.50 & 62.17 & 58.23 \\ \hline
            % Word-based (LSTM) & 36.02 & 59.38 & 47.33 \\
        %   \ \ +char+bichar  & 43.40 & 60.30 & 52.33 \\
           Char-based (LSTM) & 46.11 & 55.29 & 52.77 \\
          \ \ + bichar+softword & 50.55 & 60.11 & 56.75 \\ 
          \ \ + ExSoftword & 44.65 & 55.19 & 52.42\\
          \ \ + bichar+ExSoftword & 58.93 & 53.38 & 56.02\\
          Lattice-LSTM & 53.04 & 62.25 & 58.79 \\ 
          LR-CNN (\citeauthor{gui2019cnn}, 2019) & {57.14} &\textbf{ 66.67 } & 59.92 \\ 
        %   proposed (LSTM) & {56.99} & 61.41 & \textbf{61.24} \\  \hline %& 67.15 & 56.28 & 61.24
          SoftLexicon (LSTM) & \textbf{59.08} & {62.22} & \textbf{61.42} \\
          SoftLexicon (LSTM) + bichar & {58.12} & {64.20} & {59.81} \\
          \hline
          BERT-Tagger & 65.77 & {62.05} & 63.80 \\ % 94.56 & 95.53 & 95.04
          BERT + LSTM + CRF & 69.65 & {64.62} & 67.33 \\
%  Lattice-Transformer + BERT & 69.23 & - & - \\
  SoftLexicon (LSTM) + BERT & \textbf{70.94} & \textbf{67.02} & \textbf{70.50} \\ \hline
    \end{tabular}}
    \caption{Performance on Weibo. NE, NM and Overall denote F1 scores for named entities, nominal entities (excluding named entities) and both, respectively.}
    \label{tab:weibo}
\end{table}

\begin{table}[t]
    \centering
    \resizebox{0.95\columnwidth}{!}{
    \begin{tabular}{lcccc}
          \textbf{Models} & \textbf{P} & \textbf{R} & \textbf{F1} \\ \hline \hline
          Word-based (LSTM) & 93.72 & 93.44 & 93.58 \\
          \ \ +char+bichar  & 94.07 & 94.42 & 94.24 \\
           Char-based (LSTM) & 93.66 & 93.31 & 93.48 \\
          \ \ + bichar+softword & 94.53 & 94.29 & 94.41\\ 
          \ \ + ExSoftword & 95.29 & 94.42 & 94.85\\
          \ \ + bichar+ExSoftword & \textbf{96.14} & 94.72 & 95.43\\
          Lattice-LSTM & 94.81 & 94.11 & 94.46\\ 
          LR-CNN (\citeauthor{gui2019cnn}, 2019)& 95.37 & 94.84 & 95.11 \\ 
        %   proposed (LSTM)  & \textbf{95.97} & \textbf{94.91} & \textbf{95.43}\\  \hline
        %   proposed (LSTM)  & {95.53} & \textbf{95.64} & \textbf{95.59}\\
          SoftLexicon (LSTM)  & {95.30} & 95.77 & {95.53}\\
          SoftLexicon (LSTM) + bichar  & {95.71} & 95.77 & \textbf{95.74}\\
          \hline 
        %   BERT-Tagger & 96.12 & 95.45 & 95.78 \\ % 94.56 & 95.53 & 95.04
        BERT-Tagger & {94.87} & \textbf{96.50} & 95.68 \\
%  Lattice-Transformer + BERT & \textbf{95.97} & \textbf{96.44} & \textbf{96.21} \\
        BERT + LSTM + CRF & {95.75} & 95.28 & 95.51 \\
  SoftLexicon (LSTM) + BERT & \textbf{96.08} & {96.13} & \textbf{96.11} \\ \hline
    \end{tabular}}
    \caption{Performance on Resume.}
    \label{tab:resume}
\end{table}

% \paragraph{MSRA.} Table \ref{tab:msra} shows results on MSRA. The word-based methods were built on the automatic segmentation results generated by the segmenter trained on training data of MSRA.
% Compared methods includes the best statistical models on this data set, which leveraged rich handcrafted features \cite{chen2006chinese,zhang2006word,zhou2013chinese}, character embedding features \cite{lu2016multi}, and radical features \cite{dong2016character}. From the table, we can observe that our method also outperforms Lattice-LSTM and performs comparatively with LR-CNN. This verifies the effectiveness of our method on MSRA. 

% \paragraph{Weibo/Resume.}  Table \ref{tab:weibo} shows results obtained on Weibo NER, where NE, NM, and Overall denote the F1 scores for datasets including named entities only, nominal entities (excluding named entities) only and both, respectively. The compared state-of-the-art model \cite{peng2016improving} explored rich embedding features, cross-domain data, and semi-supervised data. From the table, we can see that our proposed method achieved considerable improvement over the compared baselines on this data set. %Another notable observation is that the bichar+ExSoftword-based model achieves considerable improvement over the bichar-softword-based model on this data set. 
% Table \ref{tab:resume} shows results on Resume. Consistent with the results obtained on
% the other three tested data sets, our proposed method significantly outperformed Lattice-LSTM and the other comparable methods on this data set. 

\paragraph{MSRA/Weibo/Resume.} Tables \ref{tab:msra}, \ref{tab:weibo} and \ref{tab:resume} show results on the MSRA, Weibo and Resume datasets, respectively. Compared methods include the best statistical models on these data set, which leveraged rich handcrafted features \cite{chen2006chinese,zhang2006word,zhou2013chinese}, character embedding features \cite{lu2016multi, peng2016improving}, radical features \cite{dong2016character}, cross-domain data, and semi-supervised data \cite{he2017unified}. From the tables, we can see that the performance of the proposed Soft-lexion method is significant better than that of Lattice-LSTM and other baseline methods on all three datasets.
% while being comparable to the LR-CNN on MSRA dataset.

\subsection{Transferability Study}
%In this experiment, we studied the applicability of our method when the sequence modeling layer was implemented with the CNN- and transformer- based architectures. In addition, we tested the inference speed of our method, compared with these implementations compared with Lattice-LSTM.

% \begin{figure}
% \centering
% \begin{minipage}{3.8cm}
% \centering
% \includegraphics[width=1\columnwidth]{length_f1.pdf}
% % \caption{World Map}
% \end{minipage}
% \begin{minipage}{3.8cm}
% \centering
% \includegraphics[width=1\columnwidth]{length_speed.pdf}
% % \caption{Concrete and Constructions}
% \end{minipage}
% \caption{F1 score and inference speed against sentence length. We use a same batch size of 1 for a fair speed comparison. }
% \label{fig:lengthtest}
% \end{figure}

\begin{table}[t]
    \centering
    \resizebox{\columnwidth}{!}{
    \begin{tabular}{lcccc}
          \multirow{1}{*}{Models} & \multicolumn{1}{c}{OntoNotes} & \multicolumn{1}{c}{MSRA} & \multicolumn{1}{c}{Weibo} & \multicolumn{1}{c}{Resume} \\  \hline \hline
          SoftLexicon (LSTM)                      & 75.64             & 93.66              & 61.42             & 95.53             \\\hline
          \makecell[l]{ExSoftword (CNN)}         & 68.11             & 90.02              & 53.93             & 94.49              \\
          SoftLexicon (CNN)                       & \textbf{74.08}    & \textbf{92.19}     &\textbf{59.65}     & \textbf{95.02}     \\ \hline
          \makecell[l]{ExSoftword (Transformer)} & 64.29             & 86.29              & 52.86             & 93.78              \\
          SoftLexicon (Transformer)               & \textbf{71.21}    & \textbf{90.48}     & \textbf{61.04}    & \textbf{94.59}     \\ \hline
    \end{tabular}}
    \caption{F1 score with different implementations of the sequence modeling layer. ExSoftword is the shorthand of Char-based+bichar+ExSoftword.}
    \label{tab:transfer}
\end{table}

% \begin{figure}
% % \centering
% \includegraphics[width=0.5\columnwidth]{length_f1.pdf}
% \includegraphics[width=0.5\columnwidth]{length_speed.pdf}
% \caption{The overall architecture of the proposed method. }
% \label{fig:lengthtest}
% \end{figure}

Table \ref{tab:transfer} shows the performance of the SoftLexicon method when implementing the sequence modeling layer with different neural architecture. From the table, we can first see that the LSTM-based architecture performed better than the CNN- and transformer- based architectures. In addition, our method with different sequence modeling layers consistently outperformed their corresponding ExSoftword baselines. %This shows that our method applies to different neural sequence modeling architectures for exploiting lexicon information.
This confirms the superiority of our method in modeling lexicon information in different neural NER models.

\subsection{Combining Pre-trained Model}
% We also conducted experiments on the four datasets to show the effectiveness of our method when the BERT pre-trained model is combined, and the results are shown in Tables \ref{tab:ontonote}$-$\ref{tab:resume}. 
We also conducted experiments on the four datasets to further verify the effectiveness of SoftLexicon in combination with pre-trained model, the results of which are shown in Tables \ref{tab:ontonote}$-$\ref{tab:resume}. In these experiments, we first use a BERT encoder to obtain the contextual representations of each sequenc, and then concatenated them into the character representations.
% The results show that the Soft-lexicon method can be properly combined with BERT, which means the lexicon-based method is not conflicted or overlapped with the pre-trained model. 
From the table, we can see that the SoftLexicon method with BERT outperforms the BERT tagger on all four datasets. These results show that the SoftLexicon method can be effectively combined with pre-trained model. Moreover, the results also verify the effectiveness of our method in utilizing lexicon information, which means it can complement the information obtained from the pre-trained model.
% verifying the effectiveness of our method in integrating lexicon information, which is shown to be complementary to the information from pre-trained model.
% is complementary to pretrained model?

\begin{table}[t]
    \centering
    \resizebox{\columnwidth}{!}{
    \begin{tabular}{lcccc}
          \multirow{1}{*}{Models} & \multicolumn{1}{c}{OntoNotes} & \multicolumn{1}{c}{MSRA} & \multicolumn{1}{c}{Weibo} & \multicolumn{1}{c}{Resume} \\  \hline \hline
          SoftLexicon (LSTM)                      & 75.64             & 93.66              & 61.42             & 95.53             \\
            \ \ - ``M" group & 75.06 & 93.09 & 58.13 & 94.72 \\ 
            % \ \ - B - E  & -    & -     &-     & -     \\ 
            \ \ - Distinction     & 70.29    & 92.08     & 54.85    & 94.30    \\ 
            \ \ - Weighted pooling & 72.57 & 92.76 & 57.72 & 95.33 \\
            \ \ - Overall weighting & 74.28 & 93.16 & 59.55 & 94.92 \\ \hline
    \end{tabular}}
    \caption{An ablation study of the proposed model.}
    \label{tab:ablation}
\end{table}

\subsection{Ablation Study}
To investigate the contribution of each component of our method, we conducted ablation experiments on all four datasets, as shown in table \ref{tab:ablation}.
% From the table, we can observe that:

(1) In Lattice-LSTM, each character receives word information only from the words that begin or end with it. Thus, the information of the words that contain the character inside is ignored. However, the SoftLexicon prevents the loss of this information by incorporating the ``Middle" group of words. In the `` - `M' group" experiment, we removed
the "Middle" group in SoftLexicon, as in Lattice-LSTM. The degradation in performance on all four datasets indicates the importance of the ``M" group of words, and confirms the advantage of our method.
% , which contains the related character inside,
% apart from beginning or ending with each character, 
% also brings word information that is beneficial to entities tagging. Such information of ``Middle" group words is ignored in Lattice-LSTM, where each character only receives lexicon information from the words that begin or end with it, i.e., it only covers the ``B" and ``E" groups.

(2) Our method proposed to draw a clear distinction between the four ``BMES" categories of matched words. To study the relative contribution of this design, we conducted experiments to remove this distinction, i.e., we simply added up all the weighted words regardless of their categories. The decline in performance verifies the significance of a clear distinction for different matched words. 

(3) We proposed two strategies for pooling the four word sets in Section \ref{sec:method}. In the ``- Weighted pooling" experiment, the weighted pooling strategy was replaced with mean-pooling, which degrades the performance. Compared with mean-pooling,
% that equally treats each word,
the weighting strategy not only succeeds in weighing different words by their significance, but also introduces the frequency information of each word in the statistical data, which is verified to be helpful.

(4) 
% We further explore the contribution of the weighting strategy in the ``- Overall weighting" experiment. 
Although existing lexicon-based methods like Lattice-LSTM also use word weighting, unlike the proposed Soft-lexion method, they fail to perform weight normalization among all the matched words. For example, Lattice-LSTM only normalizes the weights inside the ``B" group or the "E" group. In the ``- Overall weighting" experiment, we performed weight normalization inside each ``BMES" group as Lattice-LSTM does, and found the resulting performance to be degraded. This result shows that the ability to perform overall weight normalization among all matched words is also an advantage of our method. 

% These ablation results illustrate the better performance of Soft-lexicon over other models, showing the delicate design of it despite the simple architecture. 
% These ablation results illustrates the delicate design of it despite the simple architecture.

% \begin{table}[t]
%     \centering
%     \resizebox{\columnwidth}{!}{
%     \begin{tabular}{lcccc}
%           \multirow{1}{*}{Model} & \multicolumn{1}{c}{OntoNotes} & \multicolumn{1}{c}{MSRA} & \multicolumn{1}{c}{Weibo} & \multicolumn{1}{c}{Resume} \\  \hline \hline
%           proposed (LSTM)                      & 75.54             & 93.50              & 61.24             & 95.59             \\
%           proposed (LSTM) +bichar & - & - & - & - \\ \hline

%           BERT-tagger                       & -    & -     &-     & -     \\ 
%           proposed (LSTM) + BERT               & -    & -     & -    & -    \\ \hline
%     \end{tabular}}
%     \caption{F1 score of the proposed model with BERT pre-trained model incorporated, compared with the BERT tagger.}
%     \label{tab:BERT}
% \end{table}

\section{Conclusion}
In this work, we addressed the computational efficiency of utilizing word lexicons in Chinese NER. To obtain a high-performing Chinese NER system with a fast inference speed, we proposed a novel method to incorporate the lexicon information into the character representations. Experimental studies on four benchmark Chinese NER datasets reveal that our method can achieve a much faster inference speed and better performance than the compared state-of-the-art methods. 

\section*{Acknowledgements}
The authors wish to thank the anonymous reviewers for their helpful comments. This work was partially funded by China National Key R\&D Program (No. 2018YFB1005104, 2018YFC0831105, 2017YFB1002104), National Natural Science Foundation of China (No. 61976056, 61532011, 61751201), Shanghai Municipal Science and Technology Major Project (No.2018SHZDZX01), Science and Technology Commission of Shanghai Municipality Grant  (No.18DZ1201000, 16JC1420401, 17JC1420200).

% \bibliography{acl2020}
\bibliographystyle{acl_natbib}

\end{CJK}\end{document}